# Learnings from Kaggle's Forecasting Competitions

By Casper Solheim Bojer & Jens Peder Meldgaard

## Abstract
Competitions play an invaluable role in the field of forecasting, as exemplified through the recent M4 competition. The competition received attention from both academics and practitioners and sparked discussions around the representativeness of the data for business forecasting. Several competitions featuring real-life business forecasting tasks on the Kaggle platform has, however, been largely ignored by the academic community. We believe the learnings from these competitions have much to offer to the forecasting community and provide a review of the results from six Kaggle competitions. We find that most of the Kaggle datasets are characterized by higher intermittence and entropy than the M-competitions and that global ensemble models tend to outperform local single models. Furthermore, we find the strong performance of gradient boosted decision trees, increasing success of neural networks for forecasting, and a variety of techniques for adapting machine learning models to the forecasting task.

## 1   Introduction

Forecasting is concerned with accurately predicting the future and is a critical input to many planning processes in business, such as financial planning, inventory management, and capacity planning. Considerable interest has therefore been devoted in both industry and academia to the development of methods capable of accurate and reliable forecasting, with many new methods proposed each year. Forecasting competitions, where methods are compared and evaluated empirically on a variety of time series, are widely considered the standard in the forecasting community as they evaluate forecasts constructed ex-ante consistent with the real-life forecast setting (Hyndman, 2020).

Multiple forecast competitions have been held in the forecasting community during the past 50 years, with the M-competitions organized by Spyros Makridakis and Michèle Hibon gathering the most attention. The recent M4 competition attempted to take stock of new methods developed in the past 20 years and answer some questions unanswered by previous competitions. One hypothesis of the M4-competition regarded the accuracy of recent modeling advances in the machine learning (ML) community. These models are substantially more complex than conventional time series models, and the organizers predicted these would not be more accurate than simpler methods in line with previous competition findings. The findings of the M4-competition supported the claim of earlier research that combinations exhibit superior performance over single models, but the hypothesis concerning the superior performance of simple models was not supported, as the top two solutions utilized complex methods from ML (Makridakis et al., 2020b).

The relevance of the findings of the M4-competition for the domain of business forecasting has been the topic of some discussion, as some practitioners have questioned the representativeness of the competition dataset. They argue that it does not represent many of the forecasting tasks faced by business organizations (Darin & Stellwagen, 2020; Fry & Brundage, 2019). The main critique points concern the underrepresentation of high-frequency series at weekly, daily, and sub-daily levels and the lack of access to valuable information external to the time series, such as exogenous variables and the time series hierarchy. The organizers acknowledged both critique points (Makridakis et al., 2020a), and thus further research on the relative performance of methods for forecasting of higher frequency business time series with access to external information is still required. As a response to the critique, a new competition dealing with the above issues was announced in the form of the M5 competition, which will take place on the online data science platform Kaggle[1] (M Open Forecasting Center, 2019).

Several forecasting competitions addressing real-life business forecasting tasks have, however, already been completed on the Kaggle platform, but the forecasting community has largely ignored the results of these. We believe that these competitions present a learning opportunity for the forecasting community and that the findings from the competitions might foreshadow the findings of the M5 competition.

To provide an overview of what the forecasting community can learn from Kaggle's forecasting competitions, we first identified six recent and relevant forecasting competitions. We then analyzed the competition datasets in terms of their forecasting task and time series characteristics and compared them to the M3 and M4 competitions. To ensure the Kaggle solutions add value above simple and proven forecast methods, we follow up by benchmarking the Kaggle solutions. Afterward, we review the top-performing solutions of the competitions and contrast our findings with those of the M4 competition. Based on these learnings, we provide hypotheses for the findings of the upcoming M5 competition.

The rest of the paper is organized as follows. We provide background on forecasting competitions with emphasis on the recent M4 competition in Section 2, while we describe the competitions selected for review in Section 3. Following, we present an analysis of the six competition datasets in Section 4 and benchmark the competition solutions in section 5. We then proceed with the review of the top-performing solutions in the competitions in Section 6. Finally, we summarize and discuss our findings, as well as provide hypotheses for the findings of the upcoming M5 competition in Section 6 and conclude on the learnings from the Kaggle competitions in Section 7.

## 2 Background

The M-competitions have been highly influential in the forecasting community, as they focused the attention of the community on the empirical accuracy of methods rather than theoretical properties of models. Additionally, the competitions allowed anyone to participate, enabling contestants with different preferences and skillsets to use their favorite models. This allowed for a fairer comparison of methods and tapped into the diverse modeling competencies present in the forecasting community. We refer the reader to the article by Hyndman (2020) for a review of the first three M-competitions and focus our attention on the most recent competition. The M4 competition addressed feedback from the previous competitions by Makridakis et al. (2020b):

- including higher frequency data in the form of weekly, daily and hourly data,
- requesting prediction intervals to address forecast uncertainty
- emphasizing reproducibility,
- incorporating many proven methods as benchmarks,
- increasing the sample size to 100.000 time series to address concerns regarding the statistical significance of the findings.

The time series included in the competition were mainly from the business domain and were restricted to continuous time series. Besides, time series were not allowed to be intermittent or have missing values, and more than three full seasonal periods were required at each frequency (Spiliotis et al., 2020), except for the weekly time series where only 80 observations were required (Makridakis et al., 2020b).

---

[1] https://www.kaggle.com/

The findings of the competition can be divided into four main topics: i) complex vs. simple models, ii) cross-learning, iii) prediction uncertainty and iv) ensembling, see Makridakis et al. (2020b) for further details. On the topic of complex vs. simple models, the competition found that complex ML methods can outperform simple models often used for time series forecasting, as the top two solutions used a neural network and gradient boosted decision trees, respectively. It is important to note that these methods had been adapted to forecasting and thus were not out-of-the-box ML models, which performed poorly as hypothesized by the organizers. The competition also demonstrated the benefits of cross-learning, where time series patterns are learned across multiple time series. The top two performers both used models estimated on many time series, which presents a different approach than the predominant approach of one model per time series. One of the most surprising findings of the competitions concerned the remarkably accurate estimation of prediction uncertainty by the winner. The task of accurately estimating uncertainty has presented a longstanding challenge in the forecasting field, where most methods underestimate uncertainty (Fildes & Ord, 2007). Finally, the competition once again confirmed that combinations of forecasting methods, known in ML as ensembling, produced more accurate results than single methods.

Comparing the forecasting performance of ML methods and statistical methods was one of the goals of the M4 competition. However, the usefulness of these terms for categorization has been debated. Januschowski et al. (2020) argue that the distinction is unhelpful, as it does not provide a clear and objective categorization of methods. As an example, the two top methods in the M4 competition use methods associated with the ML term in the form of neural networks and gradient boosted decision trees (GBDT), as well as methods associated with the statistical term such as exponential smoothing and other classical time series forecasting methods. As an alternative, Januschowski et al. (2020) argue for the use of a more comprehensive set of classification dimensions, including global vs. local models, linear vs. non-linear, data-driven vs. model-driven, ensemble vs. single models, and interpretable vs. predictive.

Kaggle is an online data science platform that hosts data science competitions for business problems, recruiting, and academic research purposes. The platform houses a large community of data scientists from a variety of backgrounds that compete in the competitions and participate in the discussion forums by sharing knowledge and discussing potential strategies. In the business problem-focused competitions, companies provide a dataset for a prediction task of relevance and typically offer a cash prize for the top performers. In contrast to most academically hosted forecasting competitions, the Kaggle competitions provide real-time feedback on submitted predictions in the form of a publicly available leaderboard, which shows a ranked list of the contestants and their scores. Contestants are allowed to submit multiple predictions, facilitating learning, and resulting in better predictions (Athanasopoulos & Hyndman, 2011). The final competition results are based on private leaderboard performance, which is evaluated on an unseen dataset to prevent overfitting to the leaderboard.

## 3 Selection of Competitions

We initially examined the database of competitions from the online data science platform Kaggle, and only competitions focused on forecasting were kept for further consideration. Afterward, the pool of competitions was reduced by only considering competitions from 2014 and forward, as a lot has happened in the field of forecasting and ML since 2014. This reduced the pool of competitions to the following seven competitions:

- *Walmart Store Sales Forecasting*[2]
- *Rossmann Store Sales*[3]
- *Walmart Sales in Stormy Weather*[4]
- *Grupo Bimbo Inventory Demand*[5]
- *Wikipedia Web Traffic Time Series Forecasting*[6]
- *Corporación Favorita Grocery Sales Forecasting*[7]
- *Recruit Restaurant Visitor Forecasting*[8]

After a more thorough review of the datasets for the competitions, we excluded the *Grupo Bimbo Inventory Demand* competition from further review. The dataset consisted of a maximum of seven observations per time series at the weekly level with many time series having only one observation, making it unsuitable for time series forecasting.

The six competitions selected for review, along with the characteristics of their forecasting tasks, are summarized in table 1. Four of the six competitions are from the retail domain, with the remaining two being from the web traffic and restaurant domains. While the retail competitions are from the same domain, they are very different in terms of the number of time series and aggregation levels. The *Walmart Store Sales* and the *Rossmann* competitions are both at a very high aggregation level in terms of the business hierarchy with relatively few series and require forecasts for dollar sales by store/department/week and store/day, respectively. The *Corporación Favorita* competition and the *Walmart Stormy Weather* competition are on the other hand both at a very disaggregated level with forecasts of unit sales being required by product/store/day. However, they differ in the number of time series.

The remaining two competitions both feature forecasts at a disaggregate level, as daily forecasts of visits by webpage and restaurants are required, but the domains and the number of time series differ. Due to its' domain, the *Recruit Restaurant* dataset contains data on upcoming reservations, which is expected to contain useful information for the forecasting task. The *Wikipedia* dataset is a more traditional large-scale forecasting task, although it provides access to the forecast hierarchy.

---

[2] https://www.kaggle.com/c/walmart-recruiting-store-sales-forecasting
[3] https://www.kaggle.com/c/rossmann-store-sales
[4] https://www.kaggle.com/c/walmart-recruiting-sales-in-stormy-weather
[5] https://www.kaggle.com/c/grupo-bimbo-inventory-demand
[6] https://www.kaggle.com/c/web-traffic-time-series-forecasting
[7] https://www.kaggle.com/c/favorita-grocery-sales-forecasting
[8] https://www.kaggle.com/c/recruit-restaurant-visitor-forecasting

Table 1: Selected Kaggle Competitions

| Competition | Time Unit | Forecast Unit | #Observations | #Timeseries | Horizon | Accuracy Measure |
|---|---|---|---|---|---|---|
| Walmart Store Sales (2014) | Weekly | $ Sales by Department | 143 | 3331 | 1-39 | WMAE |
| Walmart Stormy Weather (2015) | Daily | Unit Sales by Product & Store | 851-1011 | 255 | 1-7 | RMSLE |
| Rossmann (2015) | Daily | $ Sales by Store | 942 | 1115 | 1-48 | RMSPE |
| Wikipedia (2017) | Daily | Views by Page and Traffic Type | 970 | ~145k | 12-42 | SMAPE |
| Corporación Favorita (2018) | Daily | Unit Sales by Product & Store | 1684 | ~210k | 1-16 | NWRMSLE |
| Recruit Restaurant (2018) | Daily | Visits by Restaurant | 478 | 821 | 1-39 | RMSLE |

The competitions considered are thus varied in terms of their characteristics but present a much more limited subset of the business forecasting tasks faced by companies than present in the M4 competition. Still, the competitions are more representative of reality in terms of the information available, as both exogenous variables and the business hierarchy is available to exploit in the creation of forecasts.

## 4 Analysis of Competition Datasets

The goal of analyzing the identified Kaggle competition datasets is to position these relative to the M3 and M4 competitions. The datasets used in Kaggle competitions depict the reality of the forecasting task of known companies, and hence we know that these are representative of particular real-world contexts. To position the Kaggle competition datasets in relation to the M3 and M4 datasets, we utilize the methodology developed by Kang et al. (2017) to represent a single time series in two-dimensional space to allow the analysis of large-scale time series datasets. Furthermore, it allows the discussion of whether the results of the M3 and M4 competitions apply in contexts similar to those seen in the identified Kaggle competitions.

### 4.1 Data Preprocessing

Since all of the identified Kaggle competitions represent a particular real-world context, the selection criteria applied to the M4 competition discussed in section 2 in regards to length and regularity of the time series are not valid. Hence, some initial preprocessing is necessary to allow extrapolation of the time series instance space. All preprocessing was performed using the R packages **data.table** (Dowle & Srinivasan, 2019) and **base** (R Core Team, 2019) and can be summarized in five steps:

1. Set NA or Negative values to zero.
2. Remove time series with all zero values.
3. If a test set is available, keep only time series present in both the training and test set.
4. Transform from irregular to regular time series by filling in missing intervals with zero.
5. Remove leading zeros.

See Appendix A for a full description of the preprocessing steps and their implications.

## 4.2 Competition Representativeness

Due to their ability to provide useful information about the M3-competition data, Kang et al. (2017) proposed a set of features *F1, F2, …, F6* that enable any time series, of any length, to be summarized as a feature vector **F** = (*F1, F2, F3, F4, F5, F6*):

1. The *spectral entropy* (*F1*), as defined by Goerg (2013), measures "forecastability".
2. The *strength of the trend* (*F2*) measures the influence of long-term changes in the mean level of the time series.
3. The *strength of seasonality* (*F3*) measures the influence of seasonal factors.
4. The *seasonal period* (*F4*) explains the length of periodic patterns.
5. The *first-order autocorrelation* (*F5*) measures the linear relationship between a time series and the one-step lagged series.
6. The *optimal box-cox transformation parameter* (*F6*) measures if the variance is approximately constant across the whole series.

To calculate the feature vectors, we use the R package **feats** (O'Hara-Wild et al., 2019), and subsequently, apply principal components for dimensionality reduction using the prcomp algorithm from the R package **stats** (R Core Team, 2019) to project them all in two-dimensional space to allow for easy visualization with the R package **ggplot2** (Wickham, 2016). A similar method is also used by Spiliotis et al. (2020) in their assessment of the representativeness of the M4 competition. In their study, they use the *seasonal period (F4)* defined by the authors of the M4 competition, namely that yearly, weekly, and daily time series have a seasonal period of one, quarterly time series have a seasonal period of four, monthly time series have a seasonal period of 12, and hourly time series have a seasonal period of 24. As such, seasonality cannot be estimated for weekly and daily series since the estimation algorithm requires a seasonal period higher than one. In our approach, we substitute the original seasonal period for weekly time series with 52 and daily time series with seven. Thus, we assume that weekly time series exhibit an annual (52 weeks) seasonality, and daily time series exhibit a weekly (seven days) seasonality, thereby enabling the estimation of the *strength of seasonality (F3)*. Furthermore, we excluded *seasonal period (F4)* from the dimensionality reduction since this has little value other than separating the time series feature vectors according to their specified seasonal period. For a full description of the implications of altering and excluding the *seasonal period (F4)*, see Appendix B.

Figure 1 depicts the resulting time series instance space for both the M3, M4, and Kaggle competitions with the density of the data illustrated in each hexbin region with low density in dark grey and high density in blue. All plots, but the M4 plot, include a light gray background illustrating the M4 time series instance space, while the M4 plot includes a light gray background of the combined time series instance space of all but the M4 competition. Figure 1 show similarities in the positioning of time series between competitions with similar aggregation levels in the business hierarchy. The *Rossmann, Recruit Restaurant,* and *Walmart Store Sales* competitions are all at a high aggregation level and have density peaks within the same region of the time series instance space. We see that the highly aggregated time series have lower degrees of *trend* and *first order autocorrelation* and higher degrees *spectral entropy* than the majority of the time series in the M4 competition. The *Corporacíon Favorita*, *Walmart Stormy Weather,* and *Wikipedia* competitions are all at low aggregation levels, but the similarity of their position in the time series instance space is not as apparent. Here, we see that the *Corporacíon Favorita* and *Walmart Stormy Weather* are very similar, and both exhibit very high degrees of *spectral entropy*, low degrees of *trend* and *first order autocorrelation*, and varying degrees of *seasonality* and *lambda*. On the contrary, the *Wikipedia* competition display significantly higher *trend* and *first order autocorrelation* than the other competitions on a low aggregation level.

We expect that the dissemblance in *entropy* between the M-competitions and the Kaggle competitions, is to some extent caused by the selection criteria that prohibit intermittency in the M-competitions since contrary to the M-competitions, all of the Kaggle competitions feature some intermittent time series. Explicitly, we find that more than 98% of the time series in the *Corporacíon Favorita, Walmart Stormy Weather, Recruit Restaurant,* and *Rossmann* competitions exhibit some degree of intermittency. Furthermore, we find that approximately 16% and 26% of the time series in the *Walmart Stores Sales* and *Wikipedia* competitions exhibit intermittency, respectively. The M3 and M4 competition cover a significant part of the time series instance space, but including more time series with higher degrees of entropy and intermittence, would improve the representativeness of the competition datasets.

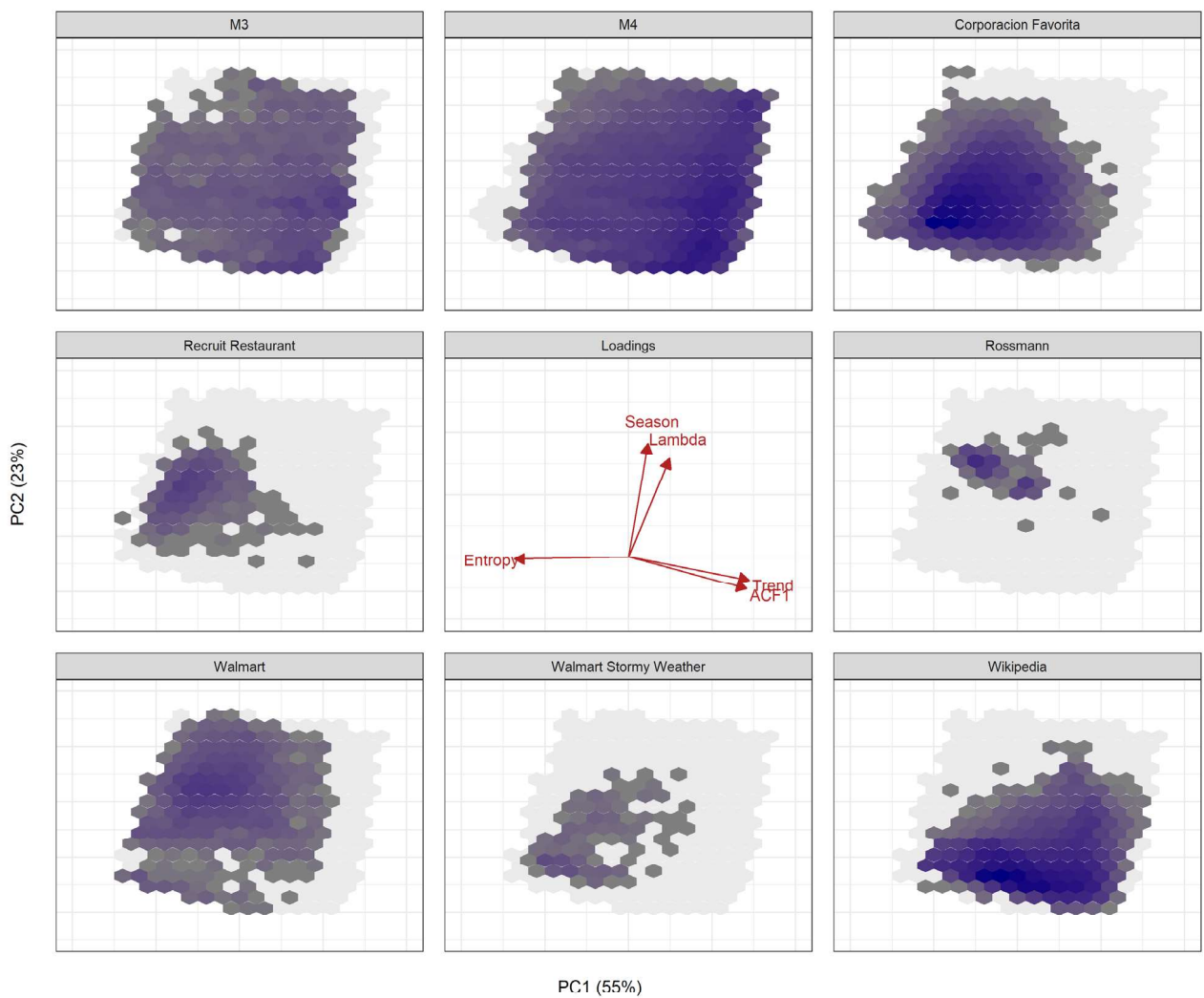

**Figure 1**: Hexbin plots of the time series instance space of M- and Kaggle competitions. The color of each hexbin illustrates the density of time series positioned in that particular field of the instance space, with blue symbolizing high density and dark grey low density. Additionally, the instance space of the M4 competition is illustrated as a light gray background on all plots, except the M4 plot, where the light gray background illustrates the combined instance space of all competition, but the M4 competition.

## 5   Benchmarking Kaggle Solutions

The fundamental premise of our article is that the learnings from top-performing solutions to the Kaggle competitions are valuable. For this to hold, the solutions should as a minimum outperform simple and proven time series forecasting methods. To verify whether this was the case, we benchmarked the solutions against two forecasting workhorses, the naïve and seasonal naïve methods. These methods are often used to identify whether a forecasting method or process adds value in e.g., forecast value added (FVA) analysis (Gilliland, 2011) or in the MASE forecast accuracy measure (Hyndman & Koehler, 2006). We choose these two methods as they are simple and robust to missing data, which is present in all of the Kaggle competitions. Other often-used benchmarking methods, such as the theta and exponential smoothing models require a time series without missing data. Therefore they have not been used, as that would require an imputation procedure to fill in missing values before forecasting.

To construct the benchmark, we produced forecasts for all the competitions using both the naïve and seasonal naïve methods. Some of the competitions required forecasts for time series that were not present in the training dataset, and we had to resort to the use of a fallback method. As a fallback, we used the mean at the next level of the forecast hierarchy to conduct some simple cross-learning. In cases where data was still missing at the next level, we proceeded up the hierarchy until data was present. For details on the specifics of the procedure used for each dataset, see Appendix C. We use relative errors to measure the performance of the solutions compared to the benchmarks, where we divide the accuracy measure of the 1st and 50th place solutions in each competition with that of the naïve and seasonal naïve methods. Table 2 shows the results of the benchmarking procedure for the better of the two benchmarking methods. Across the board, the first place solutions provide improvements above 25% over the simple benchmarks. Similarly, the 50th place solutions all beat the simple benchmarks with at least 10%. The performance improvements are particularly striking for the *Rossmann* and *Corporación Favorita* competitions, where the 1st place has reduced forecast errors by 74% and 60%, respectively. It is clear from the benchmarks that the Kaggle solutions all add value above simple time series benchmarks, and further attention is warranted.

**Table 2**: Benchmarks comparing 1st and 50th place of Kaggle competition with the Naïve and Seasonal Naïve forecasting methods.

| Competition | Method | Rel. Error 1st Place | Rel. Error 50th Place |
|---|---|---|---|
| Corporación Favorita | Seasonal Naive | 0.40 | 0.41 |
| Recruit Restaurant | Seasonal Naive | 0.73 | 0.75 |
| Rossmann | Seasonal Naive | 0.26 | 0.29 |
| Walmart Store Sales | Seasonal Naive | 0.73 | 0.89 |
| Walmart Stormy Weather | Seasonal Naive | 0.70 | 0.75 |
| Wikipedia | Naive | 0.73 | 0.82 |

# 6   Competition Review

To conduct the review, we read through the Kaggle forum posts for each of the competitions. We gathered all information on the solutions posted by contestants, including both textual descriptions and code. Solutions in the top 25 were considered for review in each competition to focus on top performers. Additionally, the forums were examined for the application of simpler methods, such as historical averages or proven forecast methods, to investigate the improvement obtained over the use of simpler methods. Table 3 shows the reported solutions for the top 25 in each competition that we identified during the review process, along with a codification of the methods used. Blank cells indicate contestants that did not describe their methods on the forums. It is evident from the table that a significant part of contestants does not report their methods, although the top performers usually do. This is a limitation of learning from the Kaggle competitions, and we discuss the implications of this in section 7.3. An analysis of table 3 also reveals another pattern: the two Walmart competitions mainly featured time series and statistical models in the top 25, whereas the four later competitions mostly feature gradient boosted decision trees and neural networks. In the following sections, we present detailed findings of the review for each of the six competitions.

Table 3: Overview of the reviewed solutions in the top 25 for each of the six Kaggle competitions. Blank cells indicate that a description of the solution was not available. The text is a codified comma separated list of the methods employed by the solutions, where multiple values indicate the use of an ensemble: **GBDT**: Gradient Boosted Decision Trees, **DTF**: Decision Tree Forests (Random Forest and Extremely Randomized Trees), **TS**: Time series methods (e.g. Exponential Smoothing, ARIMA, Kalman Filter and Moving Averages/Medians), **NN**: Neural networks, **LM**: Linear Regression, **STAT**: Other statistical methods (Polynomial regression, Projection Pursuit Regression, Unobserved Components Model, Principal Components Regression, and Singular Value Decomposition), **ML:** Other ML methods (Support Vector Machines and K-Nearest Neighbors).

| Placing | Walmart Store Sales | Walmart Stormy Weather | Rossmann | Wikipedia | Corporación Favorita | Recruit Restaurant |
|---|---|---|---|---|---|---|
| 1 | STAT, TS | STAT, LM | GBDT | NN | GBDT, NN | GBDT, NN |
| 2 | TS, STAT, DTF, ML, LM | | GBDT | GBDT, NN, LM | NN | |
| 3 | TS | GBDT | NN | NN | GBDT, NN | |
| 4 | TS | | GBDT | NN | NN | |
| 5 | TS | STAT | | STAT | GBDT, NN | GBDT, NN |
| 6 | TS | LM, DTF, ML, TS | | NN | GBDT, NN | |
| 7 | | | | NN | | GBDT |
| 8 | LM | | | TS | GBDT, NN | GBDT |
| 9 | LM | | | | | |
| 10 | GBDT, LM | | GBDT | | | GBDT |
| 11 | TS | GBDT, DTF, ML, LM | | NN, TS | | GBDT, DTF, TS |
| 12 | | | | | GBDT, NN | NN |
| 13 | TS | | | | GBDT, NN | |
| 14 | | | | NN, TS | | |
| 15 | | | | | GBDT, NN | |
| 16 | GBDT | | | NN, TS | GBDT, NN | |
| 17 | | | | | GBDT | |
| 18 | | | | | GBDT | GBDT |
| 19 | | LM | | GBDT, TS | | |
| 20 | | | | | | |
| 21 | | | | | | GBDT, NN, TS |
| 22 | | | | | | |
| 23 | | | GBDT, ML | | | GBDT, NN |
| 24 | | | | | | |
| 25 | | | | | | GBDT, NN |

## 6.1   Walmart Store Sales Forecasting (2014)

The Walmart Store Sales forecasting competition is the oldest of the competitions considered. The competition tasked contestants to provide forecasts of Weekly Sales in $ by department and store for a horizon of 1 to 39 weeks. The contestants had access to 33 months of data for the 45 stores and 81 departments, as well as metadata for the stores, holiday information, promotion indicators, weekly temperatures, fuel prices, consumer price index (CPI), and unemployment rate.

Accurate modeling of seasonality and holidays turned out to be crucial in this competition, and top-performing solutions mainly used conventional time series forecasting methods with minor tweaks. The main innovation of the winner was to learn seasonal and holiday patterns globally and use these to denoise the individual time series. The winner accomplished this using a truncated singular value decomposition (SVD) for each category of time series (department), which was used to reconstruct the individual time series. The effect of the truncation is to remove low-signal variations in the data, effectively filtering out the noise. The denoised time series were then forecasted using local forecasting methods such as STL decomposition in combination with exponential smoothing and ARIMA. Finally, forecasts from these methods were combined with an ensemble of simple forecast models such as seasonal naïve, linear trend and seasonality models, and historical averages. While all the ensembles improved his forecast accuracy, a single model from his ensemble consisting of SVD followed by STL and exponential smoothing would have been accurate enough to win the competition on its own.

An important tweak used by all contestants in the top eight was to adjust the data to lineup holidays from year-to-year, which enabled these to be modeled as part of the seasonal pattern using time series models. ML models did not fare well on their own in the competition and were used mostly as part of an ensemble, also containing time series models. One of the uses came from the second-place solution, which used a combination of ARIMA, unobserved components model, random forest, K-nearest neighbors, linear time series regression, and principal components regression for each department. As such, the models were in the middle on the global vs. local dimension. Interestingly, this relatively complicated ensemble of models did not manage to beat out the simpler first place solution.

The exogenous variables available in the competition, including temperature, fuel prices, consumer price index, unemployment, and information on markdowns, did not prove to be useful in creating accurate forecasts. While some of the top 10 contestants used them, the top two and the fourth place did not, suggesting that they added little value. Other interesting submissions to the competition include the third place due to its simplicity. The submission lined up holidays and used a weighted average of the two closest weeks from last year adjusted for the growth rate of the time series and warm days. This simple solution turned out to be only 4% worse than the best solution. As for standard time series benchmarks, the simple naïve method turned out to be a strong benchmark but was still beaten by more than 20% for all of the top 10 contestants.

## 6.2 Walmart Sales in Stormy Weather (2015)

The Walmart Sales in Stormy Weather competition featured a slightly different format than the other competitions, as the goal was to forecast the impact of extreme weather on sales. The task of the competition was to provide forecasts of Daily Unit Sales by product and store for a total of 255 time series. The format differed from the other competitions in that forecasts were not required for a future period. Instead, forecasts were required for a ±3-day window surrounding extreme weather event occurrences, which had been removed from the available data. Thus, the task was not a forecasting task in the purest sense of the word, as observations from after the forecast periods were available. To construct these forecasts, the contestants had access to 28 months of data with some extreme weather events removed for the 44 stores and 111 products, as well as extensive weather information.

The winner of the Walmart Stormy Weather competition used a variation of a common approach in retail forecasting software, which is first to estimate baseline sales and then model the deviations from the baseline using linear regression with exogenous variables. The solution used projection pursuit regression using only time as an input to estimate baseline sales per time series while taking into account trends and potentially yearly seasonality. The deviations from the baseline were modeled with a global L1-regularized linear regression model with interactions using the Vowpal Wabbit library (Vowpal Wabbit, 2020). The main difference from the typical approach in retail forecasting software is thus the use of a more complex smoother than the often-used moving average, as well as a global rather than local regression model. The winner constructed several features[9] from the exogenous variables, including modeling of weekend/weekdays, holidays, and their interactions, along with time information (year, month, day, and trend) and modeling of Black Friday (including lag and lead effects). As expected, the weather data was used in the solution in the form of indicator variables for modeling of threshold effects for precipitation and departure from normal temperatures. However, the winner mentioned in his solution write-up that using weather information did not help forecast performance by much, which is corroborated by other top-placing contestants. This finding is somewhat surprising, as the purpose of the competition was to predict the effect of extreme weather on sales and the actual weather was available instead of a forecast.

Top contestants used several other approaches, such as the use of local Gaussian Process regression by the 5th place solution using mainly date features. Ensembles of various ML models with models such as GBDT using the XGBoost algorithm (Chen & Guestrin, 2016), random forest, SVM, and linear regressions were used successfully by the 3rd, 6th, and 11th place. It is, however, interesting to note that none of these complex ensembles of models, which generally fare well in later Kaggle competitions, did better than the much simpler approach of the winner. This competition also presented the first use of XGBoost, and while placing third, its' performance was not dominating. On the other hand, conventional time series models were not used much in the reported solutions. An exception is the 6th place solution, which used time series models, such as ARIMA, as part of an ensemble. However, ARIMA did not achieve impressive performance on its own, with a performance 17% worse than the winning solution on the public leaderboard.

## 6.3 Rossmann Store Sales (2015)

The Rossmann Store Sales competition featured the rise of ensembles of global ML models, more specifically the rise of XGBoost. It was also the first competition where a neural network managed to place at the top three. The competition tasked contestants to forecast Daily Sales in $ by store for a horizon of 1 to 48 days. The contestants had access to 31 months of data for the 1115 stores, as well as metadata for the stores, promotion indicators, holiday information, weather information, and Google Trend statistics.

The winner of the competition outperformed other contestants mainly by adapting the XGBoost model to perform well on time series. This adaptation included the construction of many features using the time series and exogenous variables, as well as a trend adjustment using a ridge regression model to deal with the fact that GBDT cannot extrapolate trends. The main innovations on the feature front consisted of statistics and their rolling versions calculated at different levels of the hierarchy, for different days of the week and promotion periods. Examples include average sales by product, moving averages of sales by product, and average sales by product and promotion status. Additionally, event counters proved useful. These consist of the number of days until, into, and after an event, such as holidays or promotions. The solution also included weather information in the form of precipitation and maximum temperature together with seasonality indicators including Month, Year, Day of Month, Week of Year, and Day of Year to allow for accurate estimation of multiple seasonal effects. A key to good performance with many ML models is the appropriate selection of features and hyperparameters to maximize accuracy without overfitting the training dataset. The strategy used by many contestants was to use hold-out dataset of the same length as the forecast horizon to evaluate the quality of the model and to decide the hyperparameters and select features. The use of ensembling of multiple XGBoost models provided a performance boost of around 5% over the best single model. Variation was introduced into the ensemble by training the model on different data subsets, training models using both direct and iterated predictions, and by including different subsets of features in the models.

---

[9] In this review we use the term exogenous variables to refer to the raw information provided in the competition, and features to refer to the potentially processed inputs to the models.

Most of the top performers used ensembles of global XGBoost models to create forecasts, but a few of them did include local XGBoost models as part of their ensemble. The features used were generally similar to the winner in that they contain event counters and statistics calculated at various levels of the hierarchy. As such, the exogenous variables related to events, i.e., holidays and promotion, turned out to be essential for obtaining high performance in this competition. This likely explains the significant performance improvements obtained over the seasonal naïve benchmark. The distinguishing feature of the two best solutions is that they used rolling statistics in the form of moving averages or medians as features, thus adapting and utilizing well-known methods in the time series forecasting literature.

The 3rd place solution successfully used neural networks for the first time in the forecasting competitions on Kaggle. The neural network used was a global fully connected neural network that used the exogenous variables provided in the competition, as well as event counters for holidays and promotions. The time series aspect was handled mainly by the use of seasonality indicators. The seasonality indicators and categorical metadata were modeled using categorical embeddings, where a vector representation of the categories is learned and used by the network for prediction. The solution did not include autoregressive inputs, as is usual for neural networks in forecasting. We refer the reader to the paper posted by the contestants for further details (Guo & Berkhahn, 2016).

The highest scoring simpler method, was the 26th place that used a hybrid approach containing conventional time series models. First, a local ARIMA (with and without exogenous variables) and exponential smoothing models were used to produce forecasts. Afterward, a global XGBoost model was used to forecast their residuals based on weekday, event counters, Google Trends patterns, and weather information to capture the effects of exogenous variables not adequately modeled by the time series models. As such, traditional time series models did not fare well in the competition and were only used together with a global ML model or in the case of moving averages to construct features. The winner managed to beat the time series hybrid model by 11% and a simple benchmark consisting of the median by store, weekday, year, and promotion status by 31%, making it clear that more complex models yielded much better solutions in this competition.

## 6.4 Wikipedia Web Traffic Forecasting (2017)

The Wikipedia Web Traffic Forecasting competition took scale to another level, requiring forecasts for more than 145.000 time series. It also showcased the power of deep learning for forecasting, which won the competition and took up six places in the top eight. The competition tasked contestants to forecast daily Wikipedia page visits for a horizon of 12 to 42 days. The contestants had access to 32 months of data for the page visits, as well as metadata for the Wikipedia pages.

The winning solution presented both an elegant and accurate deep learning approach without a lot of feature engineering[10], as is typical for solutions using gradient boosted decision trees. The solution consisted of an ensemble of global recurrent neural networks with identical structures. Multiple ensembling approaches were used to reduce the variance of the predictions, as neural network predictions can prove volatile with noisy data. Three models were trained on different random seeds to counteract the randomness of the network's weight initialization. Two approaches were used to prevent sensitivity to the exact number of training iterations used. Firstly, model checkpoints were saved during the training procedure, and averages of the checkpoint predictions were used. Secondly, moving averages of neural network weights are used instead of the final weights, also known as stochastic weight averaging (SWA) (Izmailov et al., 2019). The features used in the neural networks consisted of historical page views and categorical variables such as agent, country, site as well as the day of the week. One weakness of recurrent neural networks is that they have difficulties in modeling long-term dependencies, such as is the case with yearly seasonality. The winner found a way around this by including as inputs to the model the page views from a quarter, half-year and year ago. Also, he included the autocorrelation function value at lag 365 and lag 90 to facilitate better modeling of the yearly seasonality. Series were independently scaled to facilitate cross-learning of seasonality and time dynamics, but a measure of scale in the median page views was used to allow the model to learn any potential scale-dependent patterns. A hold-out validation set was used together with an automated hyperparameter tuning algorithm using the Bayesian optimization algorithm SMAC3 (Lindauer et al., 2020) to decide on the hyperparameters of the neural networks. Interestingly, the winner reported that the final performance was relatively insensitive to hyperparameters, with the algorithm finding several models with similar performance.

---

[10] Feature engineering refers to the process of constructing features from exogenous variables or the time series itself.

The other top performers used different neural network architectures, including recurrent neural networks (RNN), convolutional neural networks (CNN), and feedforward neural networks, showcasing that several different architectures can provide similar performance. Likewise, varying degrees of feature engineering was used by the top contestants. The 4th and 6th place solutions used limited feature engineering, while the 2nd place used extensive feature engineering, including using the predictions from the various ensemble models as inputs to another model (referred to as stacking in ML). The takeaway thus seems to be that a multitude of architectures can work and that complex feature engineering is not a requirement for high performing neural network forecasts with this dataset. While neural networks did dominate the competition, another much simpler solution on the 8th place deserves mention. The contestant used a segmented approach, which included Kalman filters to predict high signal series, and a robust approach using the median of moving medians over different windows to predict low signal series. This solution was the only approach in the top that used traditional time series models, and while performing well, it was still around 6% worse than the winning solution.

### 6.5 Corporación Favorita Grocery Sales Forecasting (2018)

The Corporación Favorita Grocery Sales Forecasting competition is a good demonstration of how the Kaggle community learns from and improves on solutions to previous competitions, as both the gradient boosting approaches used in the *Rossmann* competition and the neural network approaches from the *Wikipedia* competition were utilized heavily by top contestants. The competition tasked contestants to forecast Daily Unit Sales by store and product for a horizon of 1 to 16 days for more than 210.000 time series. The contestants had access to 55 months of data for the 54 stores and 3901 products, as well as metadata for the stores and products, promotion indicators, holiday information, and oil prices.

The winner used a relatively complex ensemble of models consisting of both gradient boosting models and neural network models. One change from earlier competitions was the use of the new and significantly faster gradient boosting library LightGBM (Ke et al., 2017), which makes it easier to experiment with different features and parameters. An innovation in the solution was the training of one model per forecast horizon, rather than one model for all forecast horizons to allow the models to learn what information is useful for each horizon. While yielding good results, this approach does have the trade-off of requiring 16 models rather than 1 model. This approach was used for a LightGBM model as well as a feedforward neural network in an ensemble with two other models. These models consisted of another LightGBM model trained for all horizons and the CNN architecture that placed 6th in the *Wikipedia* competition. The features used in the feedforward neural network and the GBDT models were generally similar to the features utilized successfully in the

*Rossmann* competition. The features were mainly rolling statistics grouped by various factors such as store, item, class, and their combinations. The statistics used included measures of centrality and spread, as well as an exponential moving average.

Interestingly, the winner only used very recent data in the models, electing to drop older observations based on validation dataset performance. Thus, the final models used less than a full season of data for model fitting in the form of either one, three, or five months of data, despite multiple seasons being available. Other top placers also favored this approach, such as the 5th and 6th placers. One possible explanation of why this worked despite ignoring the yearly seasonality is the trend present in the data, as well as the short forecast horizon of only 16 days.

No simple approaches were present among the top placers of the competition, which all used similar modeling approaches, consisting of LightGBM paired with feature engineering based on rolling statistics, neural networks inspired by the successful architectures from the Wikipedia competition, or ensembles of the two. The main differences between the solutions were in the details of the feature engineering and architecture, or the validation approach used.

While the hold-out strategy has been used throughout most of the previous competitions to prevent overfitting, several contestants experimented with other validation approaches. One example was the 4th place solution, which held out a certain percentage of time series, thus relying purely on cross-learning for performance estimation. Another interesting validation approach was the use of a combination of grouped K-Fold cross-validation to estimate parameters, and time series cross-validation to estimate model performance. In the grouped K-Fold cross-validation, each time series was restricted to one fold to avoid information leakage across folds, thus also relying purely on cross-learning. The time-series cross-validation used two consecutive hold-out datasets of 16 days to estimate model performance. Despite the interesting aspect of multiple validation approaches working successfully for forecasting, the hold-out approach seems to continue to suffice, as the top three solutions used it.

## 6.6 Recruit Restaurant Visitor Forecasting (2018)

The Recruit Restaurant Visitor Forecasting competition was a confirmation of the success of previously used methods such as gradient boosted decision trees using rolling statistics, and to some degree of neural networks, in a different domain. The competition tasked contestants to forecast Daily Restaurant Visits by restaurant for a horizon of 1 to 39 days. The contestants had access to 15 months of data for the 821 restaurants, as well as metadata for the restaurants, holiday information, and reservations for restaurant visits made at different times in advance.

The winner in the competition was a team of four contestants, who used an ensemble consisting of the average of their models based on LightGBM, XGBoost, and feedforward neural networks. All of the models used features based on rolling statistics as well as lagged values of restaurant reservations, which presented the main difference from earlier competitions in different domains. Another challenge in the competition was that the test set included the "Golden Week" holiday period, which has significantly different behavior, while contestants only had access to one earlier holiday period in the training dataset. Some contestants discovered an intelligent adjustment to the data to better model these holidays with the little data available by treating holidays as Saturdays and the days prior and preceding as Fridays and Mondays, respectively. This tweak generally gave a significant performance boost when used, as evaluated after the competition by multiple top placers. Using the trick was not necessary to win the competition, as exemplified by the 1st place solution. However, it underlines the value of using domain knowledge and manual adjustments to the data to achieve the best possible performance, similar to the findings of the Walmart Store Sales competition.

The 1st and 5th place solutions used neural networks, but they were not generally as successful as in earlier competitions and were mainly used to add diversity to ensembles. The recurrent and convolutional neural network variants used successfully in the *Wikipedia* and *Corporación Favorita* competitions generally performed slightly worse than models based on boosted decision trees. The 21st, 23rd, and 25th places utilized these methods with around 2% worse accuracy than the 1st place. A potential reason for this could be the size of the dataset, which is smaller than the *Wikipedia* and *Corporación Favorita* datasets by more than a factor 100. Interestingly, a Kalman filter managed to place competitively as in the *Wikipedia* competition with a 33rd place and a performance gap of only 2.4% to the 1st place, highlighting that more traditional time series models can still be viable with exogenous variables available.

As in earlier competitions, most contestants used a hold-out dataset for validation of model performance, although both the 7th and 8th placers surprisingly managed to get high placements using a standard K-Fold validation approach, ignoring the time series nature of the data. Inspired by the innovation from the Corporación Favorita competition, a few contestants trained horizon specific models, with one model by the 1st place and the 5th place submission requiring a total of 42 models. A compromise was made by the 11th place contestant, who trained one model per week for six models in total, to still model some of the potential horizon specific effects. However, not all contestants in the top used horizon specific models, suggesting that performance improvements of the approach might not be substantial compared to the growth in the number of models.

# 7 Discussion

In terms of general modeling strategies, our review supports the findings of the M4 competitions regarding ensembles vs. single models and global vs. local models. Ensembles won all of the competitions, and thus this finding continues to hold across different domains and forecasting tasks. Global models were also used by all of the competition winners, although sometimes in combination with local models, which underlines the benefits of cross-learning for time series. The performance difference between global and local models in the conducted benchmark, e.g., in the *Walmart Store Sales competition*, suggests that access to the business hierarchy provides cross-learning benefits even higher than those found in the M4 competition. Access to exogenous variables other than the hierarchy provided substantial benefits in some competitions and very small or none in others. Information on promotions, holidays, and events proved highly useful in most of the competitions where available. Variables that would need to be forecast, namely weather and macroeconomic variables, did not seem to provide significant benefits, despite the availability of actual values.

In our review of the six competitions, we did not find one method that dominated all of the competitions. The two earliest competitions, *Walmart Store Sales and Walmart Stormy Weather*, were won by innovative use of time series and statistical methods, respectively. The four later competitions were won by non-traditional forecasting methods in the form of either gradient boosted decision trees utilizing rolling and grouped statistics, or neural networks. Additionally, there is a surprisingly similar structure in the top-performing solutions across the competitions. An interesting question is thus, why the gradient boosted decision trees or neural networks did not perform well in the first two competitions? An obvious reason is that the methods were not mature or even developed at the time of the first two competitions. Neural networks were not used for forecasting before the *Rossmann competition*, and the first successful gradient boosted decision tree algorithm in the form of XGBoost was not released until after the first Walmart competition. While XGBoost was available and used in the *Walmart Stormy Weather competition*, the method was still new, and the adaptations to the time series domain, such as rolling and grouped statistics were not used. Therefore, it is impossible to answer whether the competitions would still be won by time series and statistics methods if held today.

A better question might be, why time series and statistical methods did not perform competitively on the latest four competitions? We believe the reason is that the characteristics of the last four competition datasets are better suited to both gradient boosted decision trees and neural networks. The four latest datasets are all characterized by intermittency, and contain external information relevant to the forecasting task in the form of hierarchy information and predictive exogenous variables, such as holidays, events, promotions, and reservations. On the other hand, the *Walmart Store Sales* competition is continuous, has access to hierarchy information, and the exogenous variables contain little useful information, likely due to the high aggregation level. Altogether, this presents ideal conditions for global time series methods. The *Walmart Stormy Weather* competition has the smallest of the competition datasets and is missing business hierarchy information, which limits the opportunity for cross-learning. Furthermore, the availability of data from both before and after the required forecasting periods and the short forecast horizon makes the situation ideal for statistical smoothing methods, such as the projection pursuit regression employed by the winner. Thus, we find that for disaggregate datasets that are intermittent or contain relevant external information, ML methods outperform both time series and statistical methods. This is in line with the practical experience of forecasters at both Google (Fry & Brundage, 2019) and Amazon (Salinas et al., 2019).

As for any differences between gradient boosted decision trees and neural networks, we note that the neural networks outperformed gradient boosted decision trees in the *Wikipedia competition*, which was very large and contained no useful exogenous variables. In the other three of the latest competitions, both methods placed in the top. Neural networks are the topic of much current forecasting research and were also used by the winner of the M4 competition (Smyl, 2020). However, we are not aware of research that uses gradient boosted decision trees in combination with the strategies from the Kaggle competitions. While the second-place solution of the M4 competition used gradient boosted decision trees, it was used as a meta-learner to combine traditional time series forecasting methods (Montero-Manso et al., 2020). Therefore, further research should investigate the use of gradient boosted decision trees for forecasting, given their strong empirical performance in the competitions and several useful properties for forecasting. As the method is based on decision trees, it can learn to deal effectively with in-sample level shifts by partitioning along the time dimension. By encoding the business hierarchy using rolling and grouped statistics, it can cross-learn by partitioning on these statistics to pool information from similar time series. Furthermore, the loss function to be optimized is customizable to any function that has well-defined gradients and hessians, e.g., quantile loss as is required to forecast prediction intervals. The main weakness of gradient boosted decision trees is in extrapolating trends. However, Kaggle contestants have developed methods for dealing with this, e.g., ensembling with a linear regression that models the trend.

Throughout all six of the competitions, we find the successful use of a hold-out dataset with length equal to the forecast horizon to validate model performance and prevent overfitting. It is somewhat surprising that we do not see substantial overfitting to the validation set when it is used for multiple evaluations of ML models to select features and hyperparameters. One potential explanation for this is the public leaderboard feedback provided by the Kaggle platform, as a performance drop on the leaderboard would inform contestants that they are likely overfitting to the validation set. The approach therefore in principle corresponds to splitting the data four ways when conducting a forecast competition:

- o  A training set to estimate models,
- o  A validation set to evaluate model performance and perform model diagnostics

- o A second small validation set where only the summary performance measure is available to prevent against overfitting.
- o The final test set used for evaluating out-of-sample performance.

Further research should evaluate how this approach compares to other established forecast validation strategies such as time-series cross-validation.

## 7.1 Practical Applicability

One concern often voiced with regards to more complex methods is their practical applicability and whether the accuracy gains justify the added complexity and computational requirements (Gilliland, 2020). The solutions to the four most recent Kaggle competitions all employ data-driven ML solutions, which require the training of multiple complex models. They are thus more expensive than popular time series benchmarks in terms of cost and time. Whether this added computational cost and time is justified will ultimately depend on the cost structure of the decisions, which the forecasts are used to support, and no blanket statements can be made. Gilliland (2020) argues that the accuracy improvements obtained by ML methods in the M4 competition are not substantial enough to warrant the use of ML in practice when considering their complexity and costs. In our review, we find that the top solutions generally provide considerable improvements over simple benchmarks such as the seasonal naïve method on the order of 20% to 74%. As such, the use of more complex methods that effectively use the business hierarchy and exogenous variables should warrant serious consideration for daily and weekly business forecasting tasks. It is, however, unlikely that the winning solutions achieve the best trade-off between accuracy and complexity. The use of ensembling by the winners in the *Walmart Store Sales* and *Rossmann* competitions improved accuracy slightly. Nevertheless, their best single model would likely have proved more practical for operational use in terms of both model management complexity and computational costs.

Labor costs related to operating the forecasting system, such as data cleaning and forecast adjustments, should also be considered in this discussion, as noted by Januschowski et al. (2020). As an example, it is common in the retail sector to use simple time series methods and rely on human judgment to estimate the impact of promotions and adjust the forecasts, as simple methods are not able to model this accurately (Fildes et al., 2019). Simple methods have lower computation costs, but they also require significant human labor, whereas more complex methods using promotion information might only require labor for exception management, e.g. in the case of new price points or promotion strategies. Further research should look into evaluating this trade-off dimension e.g., by investigating labor usage in real-life forecasting systems using simple and ML methods.

## 7.2 M5 Hypotheses

The upcoming M5 competition features a hierarchical dataset from the retail domain generously supplied by Walmart. The competition will require forecasts for more than 40.000 daily time series at the store and product level, and provide contestants with information on prices, promotions, events, and product hierarchy. As such, the forecasting task is very similar to that of the Corporación Favorita competition. The main difference from the reviewed competition is the evaluation of prediction uncertainty in addition to prediction accuracy. Based on the learnings from our review, we provide the following hypotheses:

- The instance space representation of the time series in the M5 competition will resemble that of the Corporación Favorita competition, meaning that entropy is higher and trend and first-order autocorrelation is lower than time series in previous M-competitions.
- The winning method will utilize cross-learning, and global and hybrid models will dominate local models.

- Access to hierarchy information will increase the performance gap between local models and models using cross-learning compared to the M4 competition.
- GBDT using feature engineering based on e.g., rolling statistics and neural networks will both perform well in the competition and outperform existing time series benchmarks in terms of both accuracy and uncertainty.
- To provide prediction intervals, GBDT and neural networks will be adapted by using custom loss functions such as quantile loss, or by adapting the training procedure/architecture to output distributions, which has been the topic of much recent research (see e.g, Duan et al. (2020) for GBDT and Salinas et al. (2019) for neural networks).
- Ensembles of methods will continue to take up the top slots, as is consistent with the findings from all Kaggle and M-competitions. We expect these ensembles to contain both neural networks and GBDT, potentially in combination with other methods.
- Hold-out datasets or time series cross-validation will be used by top placers to avoid overfitting.
- Using exogenous variables such as prices, promotions, holidays, and other events will provide improvements to forecast accuracy, in line with previous retail research (Fildes et al., 2019) and the Kaggle competitions.
- Contestants will develop innovative strategies to tackle the challenge of hierarchical forecasting, and we expect new neural network architectures and GBDT strategies to utilize this information optimally.

## 7.3 Limitations

In contrast to the academic forecasting competitions, Kaggle's focus on providing solutions to real-life forecasting tasks has a downside, in that it provides some limitations to what researchers can infer from the competitions. Lack of access to the test set after the competition has ended means it is not possible to test for significant differences between the performances of solutions or to evaluate performance using alternative error measures. Furthermore, it is not possible to analyze the performance of different solutions on various subsets of the dataset to improve our understanding of the strengths and weaknesses of various methods.

Although Kaggle encourages the sharing of solutions, contestants are not required to share their solution or code publicly, and the lack of publicly-shared solutions has implications for the validity of our review. It is a possibility that the use of methods such as linear regression or local time series models in the non-reported solutions in the top 25 would change our results. However, we find it highly unlikely that local time series would have been able to perform competitively in the four latest competitions, which we base on the intermittency and influence of exogenous variables in the datasets. The results of our benchmarking also supports this. We also find the presence of a systematic reporting bias caused by differences in willingness to share for different solution methods unlikely. Despite these weaknesses, we still believe much can be learned by focusing on the patterns of what worked across the competitions and relating the findings to dataset characteristics. Further research should subject our hypotheses to testing on a variety of datasets, and the upcoming M5 competition will surely serve as a great initial testing ground.

## 8 Conclusions

Based on our analysis and review of the six recent Kaggle forecasting competitions, we believe that the forecast community has a lot to learn from the Kaggle community in terms of forecasting daily and weekly business time series.

In our analysis, we find that the M4 competition dataset contains time series similar to those of the Kaggle competitions, although time series with these characteristics are underrepresented in the M4 competition dataset. Furthermore, the Kaggle datasets differ from the M4 competition in that they provide access to external information, e.g., exogenous variables or business hierarchy, which led to significant improvements in forecast accuracy.

Similar to the findings from the M4 competition, we find that global ensemble models outperform local single models. In contrast to the M4 and the two earlier Kaggle competitions, conventional time series and statistical methods were significantly outperformed by machine learning methods in the four latest Kaggle competitions. We believe that this can be attributed to the machine learning methods' utilization of external information to cross-learn and model the effect of exogenous factors. Additionally, we find a similarity between top solutions in the Kaggle competitions and the top two solutions in the M4 competition, which relied on either gradient boosted decision trees or neural networks. However, to obtain the performance benefits from machine learning methods, several adaptations to the machine learning methods and their validation strategies must be adopted.

We strongly encourage the forecast community to learn from the machine learning strategies for time series forecasting and to participate in their further development. The M5 competition presents an ideal opportunity for this, as the forecasting task and dataset bears high similarity to some of the Kaggle competitions reviewed in this paper. Therefore, we believe that the learnings from the Kaggle competitions discussed in this paper will foreshadow the results of the M5 competition.

# Appendix A: Preprocessing of Kaggle Competition Datasets

Several pre-defined filters were applied beforehand to achieve some desired characteristics, relating mainly to the length of M4 series and their proportions per frequency and domain (Spiliotis et al., 2019)[11]. For instance, series with less than 10 observations or three periods were excluded as were series with missing values (Spiliotis et al., 2019)[12]. On the contrary, no pre-defined filters were applied to the time series in the Kaggle competition data sets since, and hence, preprocessing was necessary to render the series fit for analysis. All preprocessing was performed using the R packages **data.table**[13] and **base**[14] and can be summarized five steps:

1. Set NA or Negative values to zero.
2. Remove time series with all zero values.
3. If a test set is available, keep only groups that occur in both the training and test set
4. Transform from irregular to regular time series.
5. Remove leading and trailing zeros

Some of the Kaggle datasets contained NA or negative values, and with no information on how to address these, it was decided to replace them with zeros, thereby adhering to the approach used in top-scoring submissions. Next, a check was adopted to see if all values of the time series were zero, and if so, the time series were excluded from further analysis. Removing all zero series was especially important in the *Walmart Stormy Weather* competition that initially included 4,995 time series, of which 4,740 were all zero, resulting in just 255 time series to be kept.

In all competitions, except *Recruit Restaurant* and *Wikipedia*, a test set including groupings and features for the evaluation period was available. Thus, to limit the scope of the analysis, it was decided to subset the training set to only include groupings that were also present in the test set. Here, the *Walmart Stormy Weather* competition was a particular case since the objective of this competition was to predict unit sales around +- 3 days surrounding major weather events, which in the competition was defined as any day where more than an inch of rain or two inches of snow was observed. Contrary to the other Kaggle competitions, where test sets chronologically followed the training period, the weather events occurred sporadically throughout the training period and were of varying length. Therefore, unit sales were, in the training set, set to NA on all Products/Stores/Dates present in the test set and then imputed using an interpolation-based seasonal-split imputation algorithm from the R package **imputeTS**[15]. To ensure the correctness of the imputations, all the imputed series were visualized and manually inspected using the R package **ggplot2**[16].

Beyond the group, all Kaggle competitions were organized according to an index, i.e., a date. Thus, it was possible to analyze the regularity of the time series, i.e., if the interval between observations was consistent throughout the time series. The analysis showed some irregularities that were mitigated by filling in missing dates and setting the forecast unit to zero. As an example, observations were missing in the *Rossmann* competition when stores were closed. Here, the most consistent irregularity was the distinction between stores that are closed on Sundays and those that are not, which meant that the length of a week could be either six or seven days. Thus, missing Sundays were filled in with unit sales set to zero to rectify the irregularity, thereby providing a consistent week length of 7 days. Lastly, multiple time series in the Kaggle competitions either started with several consecutive zeros, which were removed.

---

[11] Are forecasting competitions data representative of the reality?
[12] Are forecasting competitions data representative of the reality?
[13] https://cran.r-project.org/web/packages/data.table/index.html
[14] https://stat.ethz.ch/R-manual/R-devel/library/base/html/base-package.html
[15] https://cran.r-project.org/web/packages/imputeTS/index.html
[16] https://cran.r-project.org/web/packages/ggplot2/index.html

# 9 Appendix B: Altering and Excluding *Seasonal Period (F4)*.

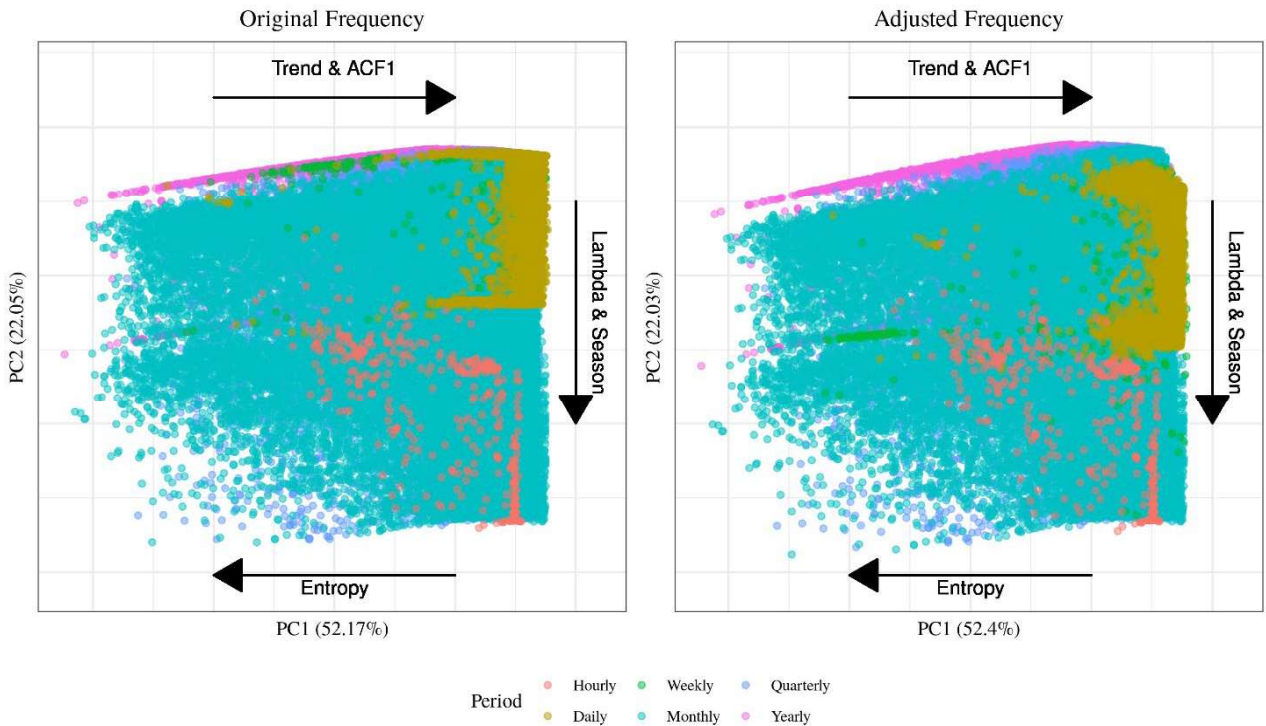

*Figure 1: M4 features with original and adjusted frequency.*

# Appendix C: Details of Benchmarking Procedure